\pdfoutput=1

\documentclass[11pt]{article}

\usepackage[final]{ACL2023}

\usepackage{times}
\usepackage{latexsym}

\usepackage[T1]{fontenc}

\usepackage[utf8]{inputenc}
\usepackage{enumerate}

\usepackage{microtype}

\usepackage{inconsolata}

\usepackage{enumitem}
\usepackage{graphicx}
\usepackage{booktabs}
\usepackage{multirow}
\usepackage{amsmath}
\usepackage{fontawesome5}

\title{\textsc{DynamicQA}: Tracing Internal Knowledge Conflicts in Language Models}

\newcommand\blfootnote[1]{%
  \begingroup
  \renewcommand\thefootnote{}\footnote{#1}%
  \addtocounter{footnote}{-1}%
  \endgroup
}

\author{Sara Vera Marjanović\textsuperscript{*} \quad \bf{Haeun Yu\textsuperscript{*}} \quad \bf{Pepa Atanasova}  \\ \bf{Maria Maistro} \quad \bf{Christina Lioma} \quad \bf{Isabelle Augenstein} \\   Department of Computer Science \\ 
University of Copenhagen\\ 
\texttt{\{savema, hayu\}@di.ku.dk}}

\begin{document}
\maketitle
\begin{abstract}

Knowledge-intensive language understanding tasks require Language Models (LMs) to integrate relevant context, mitigating their inherent weaknesses, such as incomplete or outdated knowledge. However, conflicting knowledge can be present in the LM's parameters, termed intra-memory conflict, which can affect a model's propensity to accept contextual knowledge. 
To study the effect of intra-memory conflict on an LM's ability to accept relevant context, we utilize two knowledge conflict measures and a novel dataset containing inherently conflicting data, \textsc{DynamicQA}. This dataset includes facts with a \textit{temporal} dynamic nature where facts can change over time and \textit{disputable} dynamic facts, which can change depending on the viewpoint.
\textsc{DynamicQA} is the first to include real-world knowledge conflicts and provide context to study the link between the different types of knowledge conflicts. We also evaluate several measures on their ability to reflect the presence of intra-memory conflict: semantic entropy and a novel \textit{coherent persuasion score}. 
With our extensive experiments, we verify that LMs exhibit a greater degree of intra-memory conflict with dynamic facts compared to facts that have a single truth value. Furthermore, we reveal that facts with intra-memory conflict are harder to update with context, suggesting that retrieval-augmented generation will struggle with the most commonly adapted facts.

\end{abstract}

    \faGithub 
    \hspace{2mm}\href{https://github.com/copenlu/dynamicqa}{copenlu/dynamicqa} 
    \includegraphics[width=0.35cm]{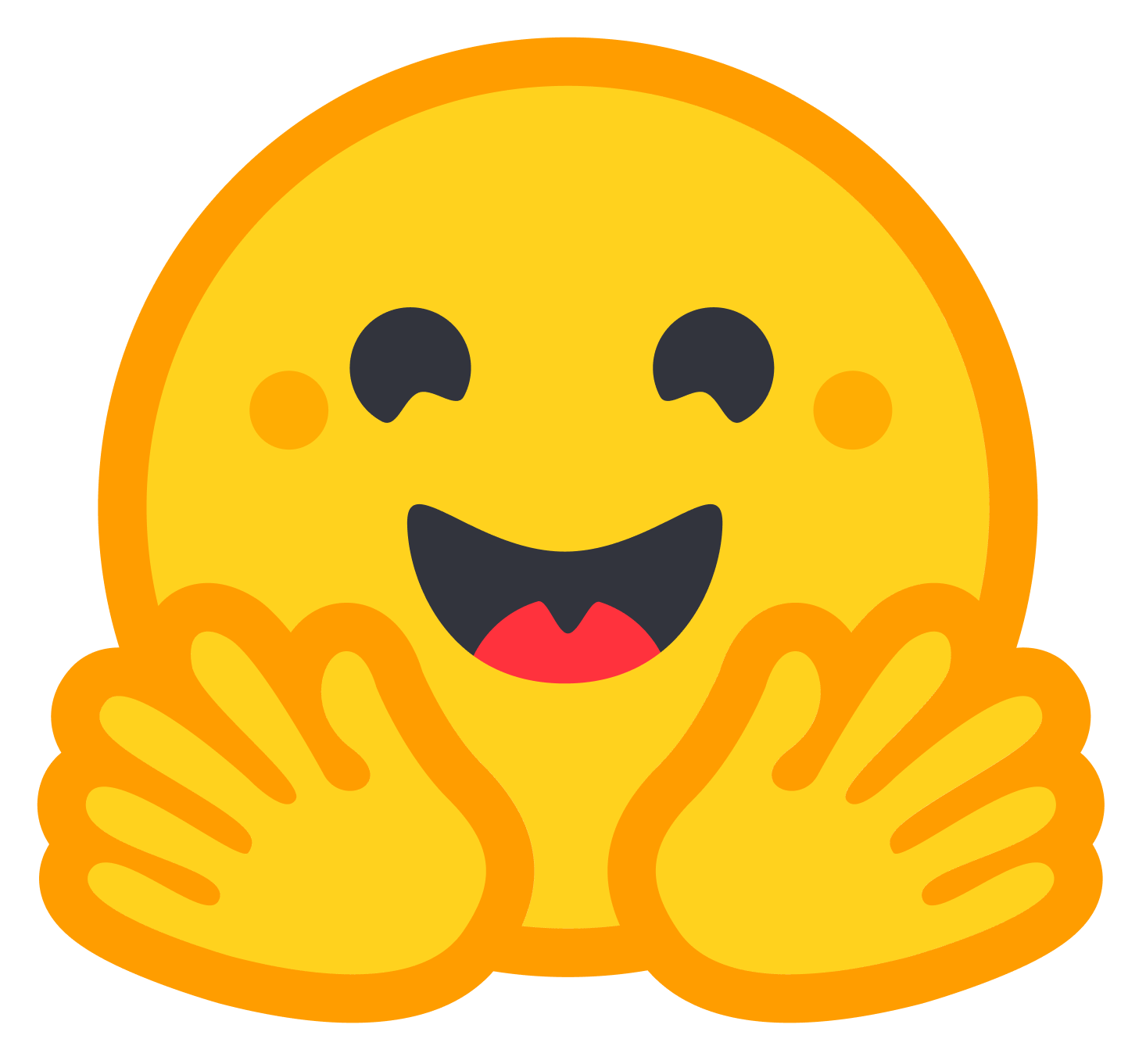}\hspace{1mm}\href{https://huggingface.co/datasets/copenlu/dynamicqa}{copenlu/dynamicqa}
    \vspace{0.5em}

\section{Introduction}
\blfootnote{\textsuperscript{*} Equal contribution. Corresponding authors.}

\begin{figure*}
    \centering
    \centerline{\includegraphics[width=1\textwidth]{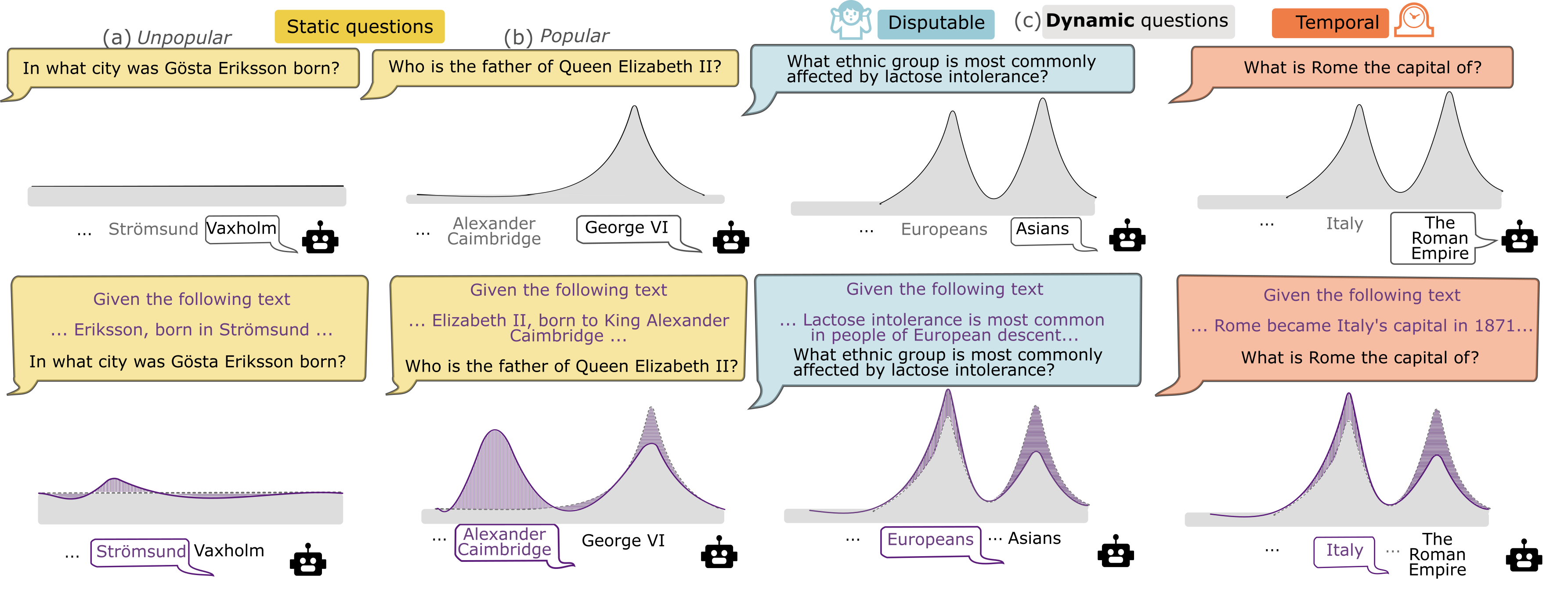}}
    \caption{We present examples of our dataset here, consisting of static, temporal, and disputable facts. We show how model output distribution can vary due to the popularity (a,b) and dynamicity (c) of facts. Fact dynamicity (c) causes intra-memory conflicts between the different fact representations seen during pretraining. In the bottom row, we show the change in output probability (purple area) that the context must enact on the initial output distribution (dotted line) to force a new output distribution (purple line).}
    \label{fig:triplets}
\end{figure*}

Language models (LMs) have been useful in a variety of downstream applications from summarization to fact-checking, often relying on the factual knowledge memorized during pre-training and stored in their parameters, known as \textbf{parametric knowledge} \cite{yu2024revealingparametricknowledgelanguage}. However, this internal memory is not infallible; it may contain misinformation, biases or simply outdated data, causing LMs to produce factually incorrect output, occasionally termed `hallucinations'\cite{huang2023survey}. Furthermore, conflicting representations of a fact may exist within training data, given conflicting viewpoints (`disputable facts') or temporal changes (`temporal facts') (See Figure \ref{fig:triplets}). We refer to this superposition of fact representation as the fact's \textit{dynamicity}. In contrast, a \textit{static} fact has only one possible representation. This can lead to \textbf{intra-memory conflicts}, which can contribute to LM's uncertainty and instability in factual recall. This intra-memory conflict can be likened to a semantic expression of aleatoric uncertainty, noise inherent in training data \cite{kendallgall2017}.

One way to guide LMs to reliable answers is via Retrieval-Augmented Generation (RAG) \cite{gu2020-Realm}, which provides additional contextual knowledge, known as \textbf{non-parametric knowledge}, to a language model.
However, this context may conflict with the model's parametric knowledge, resulting in a \textbf{context-memory conflict}.
Initial research has found a tendency of models to over-rely on parametric knowledge in the face of these knowledge conflicts \cite{longpre-etal-2021-entity, Chen2022RichKS}, thus ignoring the retrieved context. While fact popularity has been known to reduce context utilization \cite{mallen-etal-2023-trust}, little is known of the impact of internal memory conflicts. We explore this issue for the first time by looking at the effect of fact dynamicity on context utilization.

In this paper, we investigate the effect of intra-memory conflicts on context adaptation using a novel dataset \textsc{DynamicQA}. This dataset contains inherently conflicting temporal and disputable facts, paired with realistic retrieved contexts in which they occur. Furthermore, we manipulate these contexts to include similar, competing objects. Using \textsc{DynamicQA}, we study the interaction of intra-memory and context-memory conflicts by adapting two existing conflict measures. First, we approximate an LM's fact dynamicity ground truth via the number of recorded representations for a fact in Wikidata, and sample the LM's expressed intra-memory conflict using semantic entropy, which captures the semantic variation present in parametric memory \cite{kuhn2023semantic}. Furthermore, we introduce a coherent persuasion score, based on \citet{du2024context} to approximate an LM's semantic shift in output distribution given competing context.

With our framework, we show the effect of intra-memory conflict on dynamic fact question answering. We find that manipulated contexts of static facts and facts with low dynamicity show the greatest persuasiveness for LMs, despite their limited variance in the training dataset and, presumably, low likelihood to change. Furthermore, we find that semantic entropy 
alone does not reflect an instance's likelihood for persuasion given additional context, suggesting that retrieval-augmented generation language models struggle in low-certainty domains. These results underline the need for new measures of intra-memory conflict and the need for other indicators of context use in RAGs \cite{ni2024llmsneedretrievalaugmentation}.

\section{Related Work}

\paragraph{Knowledge Conflicts}

\citet{xu2024knowledge} provide a survey for knowledge conflicts in LMs; we focus specifically on intra-memory conflicts and their interplay with context-memory conflicts. \citet{longpre-etal-2021-entity} first showed that LMs ignore context that contradicts their parametric memory, generating facts not stated in the input, termed `hallucinations'. Further work either disentangles the two forms of knowledge \cite{neeman-etal-2023-disentqa} or finds other methods to resolve hallucination \cite{song-etal-2024-entity}. Other studies investigate the effect of a fact's frequency in the training data \cite{yu-etal-2023-characterizing} or the quality of the presented counter-memory \cite{xie2024sloth, wan2024evidence} on LM hallucination. Furthermore, some works find that LMs are swayed by convincing misinformation at inference-time \cite{pan2023risk, xu2024earth, wan2024evidence}. These findings suggest some pattern to a model's proclivity to ignore context. However, in contrast to these existing works, our work \textit{considers realistic, natural examples of knowledge conflict} as they occur in the real world and \textit{how different types of knowledge conflicts are connected}.

\paragraph{Measures for Knowledge Conflicts}

There have been several approaches to quantify or approximate the degree of a knowledge conflict; however, these works only focus on context-memory conflicts. \citet{pezeshkpour2023measuring} focus on the entropy of the LM's probability distribution for an answer; the difference in entropy depending on the presence or absence of additional context indicates the LM's prior knowledge on a certain fact. On the other hand, \citet{du2024context} propose a susceptibility and a persuasion score to investigate the LM when the given context contradicts the LM's parametric knowledge, based on the first token probabilities. The susceptibility score of an entity shows how easy it is to shift an LM's probability distribution of an answer regarding an entity. The persuasion score of a context represents how effective the context is at changing an LM's probability distribution of an answer. Given known issues in the reliance on first-token probabilities \cite{wang2024firsttoken}, we \textit{reformulate this score with consideration of semantic consistency} in \S\ref{sec:2:context}. Furthermore, given the lack of measures for intra-memory conflict, we \textit{assess semantic entropy \cite{kuhn2023semantic} as an indicator of intra-memory conflict \cite{gao-etal-2024-spuq}}.

\paragraph{Sources of Factual Errors} We identify three causes of model factual error: fact popularity, temporality, and disputability. Fact popularity approximates the prevalence of a fact in training data and has been shown to affect model performance \cite{mallen-etal-2023-trust} and a model's likelihood to be swayed by context (`susceptibility score', \citet{du2024context}). In contrast, temporality and disputability reflect a fact's entropy within the training and test data. For example, it is well-established that the passing of time can lead to outdated models, stressing the need for evaluation of models for a specific time-frame \cite{margatina2023dynamic, fierro2024mulan} and updating of specific temporal facts in LMs \cite{zhang2024mitigating,jang2023temporalwiki}. Similarly, \citet{xu2024knowledge} argue that context-memory knowledge conflicts arise due to either temporal misalignment or misinformation pollution of the training dataset. In a contemporary study, \citet{fierro2024mulan} show that facts with dynamic (or in their terms `mutable') relations show different patterns in model confidence, knowledge representation, and contextual alignment. All of these studies look at only one source of factual error at a time and do not assess in the context of knowledge conflicts. In our work, we 
\textit{introduce disputability} as a contributor to factual error, 
study \textit{multiple sources of factual error concurrently} (temporality, disputability, and popularity), 
\textit{assess their interaction}, and 
\textit{provide real-world proxy measures} for all three sources.

\section{\textsc{DynamicQA}}\label{sec:dataset}

To evaluate the intra-memory conflicts, we present a dataset of 11,378 question-answer pairs featuring facts with varying levels of dynamicity (see Figure \ref{fig:triplets}). The dynamicity of a fact is difficult to determine via the pretraining dataset, so we use easily measured proxy scores as a reflection of the property. The question and answer pairs, alongside their proxy scores, are sourced from Wikidata and Wikipedia and have not previously been employed in similar datasets nor for approximating the degree of knowledge conflicts in general. We approximate temporality via the number of edits (\S\ref{sec:Temporality}) and disputability via the number of reversions (\S\ref{sec:Disputability}). For each question and answer pair, we obtain answer-specific context snippets from Wikipedia, alongside their popularity scores, estimated from Wikipedia page views. For static and temporal facts, we also create `counter-memory' from these snippets using similar object replacements, to simulate realistic knowledge conflict scenarios.

\subsection{Temporality}\label{sec:Temporality}
We estimate the \textbf{temporality} of a triplet via the number of edits on Wikidata for the given subject and relation as a proxy. We initialize our dataset from PopularQA \cite{mallen-etal-2023-trust}, which is a collection of 14k question-answer pairs sampled from Wikipedia, with a \textbf{popularity} score determined by the monthly Wikipedia page views of the subject and object of the triplets. This popularity score is often used as a proxy for the triplet's prevalence in the unsearchable pre-training corpora of models \cite{mallen-etal-2023-trust, fierro2024mulan}. Given the sampling method, this dataset has a long-tailed distribution, meaning most triplets have very low popularity and low temporality. We identify relevant snippets from the triplet's subject current Wikipedia page that also mention the triplet's object and use them as context to provide to the LM. We discard 2k pairs where we cannot find the intended object and relation mentioned on the subject's Wikipedia page. To construct counter-memory contexts, we replace the object in context with a replacement token. We identify relevant replacement entities using the most similar entity for the original object as identified using the Wembedder tool \cite{nielsen2017wembedder}.
All facts with more than 1 edit are labelled as `temporal' facts, though they vary in their degree of temporality (with a maximum score of 23 edits). This leaves us with 2495 questions (Table \ref{tab:per_results}). We randomly subsample 2500 of the remaining `static' questions (triplets with 0 edits on Wikidata).

\subsection{Disputability}\label{sec:Disputability}
\textbf{Disputable} facts are facts that differ depending on the viewpoint. To construct the question and answer pairs, we utilize the collection of Wikipedia articles that are regarded as controversial.\footnote{\url{https://en.wikipedia.org/wiki/Wikipedia:List\_of\_controversial\_issues}} Controversy, by definition, is inherently disputable since it arises from the existence of multiple viewpoints, each supported by sufficient evidence \cite{wikicontroversystudy, controversy2}. We are the first to convert the controversies that appear in the Wikipedia edit history into a QA dataset for LMs.

To identify disputable facts from the controversial articles, we look for reversions in the edit logs \cite{editwars}. Among the edit logs $[e_0,\dots, e_{l-1},e_l, e_{l+1},\dots,e_j]$ on an article, two consecutive logs, $e_l$ and $e_{l+1}$ are selected as a pair of reverted edit logs if the texts of $e_{l-1}$ and $e_{l+1}$ are identical. Then, by measuring the edit distance between $e_{l}$ and $e_{l+1}$, we select a pair of words from $e_{l}$ and $e_{l+1}$ if replacing the word in $e_{l}$ with the word in $e_{l+1}$ yields the same text as $e_{l+1}$.

Since a reverted edit does not necessarily imply the disputability of a fact, we apply additional rules to filter out cases of vandalism, paraphrasing, or synonyms. For vandalism, we remove the pair if one of the users involved in that edit did not disclose their identity, for example, anonymous users or if the user ID is an IP address. For the pairs that are synonyms or paraphrasing, we feed them to a semantic similarity model (\S\ref{app:implementation}) and remove the pairs whose similarity scores are bigger than 0.98. The selected pairs of reverted edits and the text snippets they are located in serve correspondingly as answers and contexts in our dataset.

Questions are generated using an LM\footnote{\url{https://huggingface.co/meta-llama/Meta-Llama-3-8B-Instruct}}(\S\ref{app:qgen}) by providing the context $e_l$ and the corresponding ground-truth answer within the prompt (see Table~\ref{tab:prompt_qgen} in the Appendix). After obtaining the question and answer pair with the context, three annotators manually annotated the dataset to ensure its quality. Two annotators annotated each data point, and conflicts were resolved by the third annotator. We obtain a Krippendorf's alpha of 0.44 and provide further details in Appendix \ref{app:implementation}. For the real-world proxy score of disputability, we count the number of reversions on $(e_l, e_{l+1})$. As a result, we obtain 694 questions with two possible disputable answers and their corresponding real-world proxy scores (Table \ref{tab:per_results}). 

\section{Measuring Knowledge Conflicts}\label{sec:measuring}

In the pretraining stage, LMs learn a conditional distribution to predict the next tokens given a preceding context; in this process, they also obtain world knowledge into their parametric memory \cite{chang2024largelanguagemodelsacquire}. For example, in Figure \ref{fig:triplets}, given a question (e.g., `Who is the father of Queen Elizabeth II?'), the conditional probability of the correct answer `George IV' is expected to be greater than similar objects, owing to the frequent co-occurrence of the subject, object and relation in the training corpus. One important element impacting fact acquisition and recall is the frequency of the fact in the training data. It can be assumed that high frequency (often approximated via fact popularity \cite{mallen-etal-2023-trust})  leads to a greater strength of association between the subject and the object, resulting in greater probability being attributed to the relevant tokens (Figure \ref{fig:triplets}b). However, as information evolves over time, multiple different representations of a factual triplet may appear in a training corpus; meaning, given a subject and relation, there can exist a superposition of object representations \cite{Shanahan_McDonell_Reynolds_2023} depending on the temporal and situational context. Rather than a singular object being assigned a high probability, for such facts, we expect several competing answers in the output distribution (Figure \ref{fig:triplets}c).

To evaluate this hypothesis, we expect to see two properties: (1) higher entropy in the output distribution of dynamic facts and (2) greater change in the output distribution to update dynamic facts. To verify (1), we look at the semantic entropy of output distributions (\S\ref{sec:2:intra}). To investigate (2), we analyse the amount of effort required to shift the output distribution using both loss and our novel Coherent Persuasion score (\S\ref{sec:2:context}). 
Both measures look at semantic changes in LM output, thereby limiting confounds due to syntactic inconsistencies and other issues arising from over-reliance on first-token probabilities \cite{wang2024firsttoken}.

\subsection{Preliminaries}

A dataset $D=[(c_1, q_1, y_1), \dots, (c_N, q_N, y_N)]$ consists of $N$ tuple instances containing: an answer-containing context $c$, a question $q$, and a ground-truth answer $a$. For the $i$-th instance, an LM $f$ outputs an answer $y_i=[y_i^1, \dots, y_i^h, \dots, y_i^H]$, consisting of $H$ tokens given an input $x_i$. An input $x_i$ here contains a universal prompt $P$ describing the QA task with either $q_i$ only ($x_{i;q}=[P;q_i]$) or with $q_i$ and the context $c_i$ ($x_{i;q,c}=[P;q_i,c_i]$). Given a contextless input ($x_{i;q}$), a model outputs $y_{i;q}$ (i.e. the parametric knowledge). Given  context ($x_{i;q,c}$), we denote the model output as $y_{i;q,c}$. 

\subsection{Semantic Entropy}\label{sec:2:intra}

Entropy allows us to measure the amount of information within a distribution and is a common measure for model uncertainty. Both lack of information (Figure \ref{fig:triplets}a) and competing information (Figure \ref{fig:triplets}c) can contribute to uncertainty and lead to more uniform output distributions. We propose semantic entropy as an estimator of intra-memory conflict, given its use of high-temperature sampling to elicit diverse answers contained within parametric memory. We validate this measure through the lens of a fact's dynamicity (defined in \S \ref{sec:dataset}). 

To do so, we use \citet{kuhn2023semantic}'s approach in our simulated knowledge conflict setting. Given any input $x_i$, which contains a universal prompt and a question $q_i$ (and optional context $c_i$), we generate $K$ model outputs $Y=[y_{i,1}, \dots, y_{i,K}]$. Next, we group the generated answers according to their semantic similarity. The semantic similarity between two sampled answers is calculated using a DeBERTA Natural Language Inference (NLI) model.\footnote{\url{https://huggingface.co/sentence-transformers/all-MiniLM-L6-v2}} Details about answer generation and semantic grouping can be found in Appendix \ref{app:implementation}. The grouping according to the semantic similarity results in the $V$ groups $G=[g_1, g_2, \dots, g_v, \dots, g_V]$, where $1 \leq V \leq K$.

The semantic entropy is then estimated by the entropy between the semantic sets. First, we obtain $p(g_v|x_i)$, the conditional token probabilities output by the model generating the answers in $g_v$ given the input $x_i$ via Equation \ref{eq:genprob}.
\begin{equation}~\label{eq:genprob}
\begin{split}
    p(g_v|x_i) = \sum_{y_{i,k}\in g_v} p(y_{i,k}|x_i)
    \\
    = \sum_{y_{i,k}\in g_v} \prod_{h} p(y_i^h|y_i^{<h},x_i)
\end{split}
\end{equation}
In this case, $h$ refers to the intermediate tokens in the entire output length. With the probabilities of the separate groups, we approximate $SE(x_i)$, the overall expectation of the semantic entropy of $x_i$, using Monte Carlo integration over the groups:
\begin{equation}~\label{eq:su}
    SE(x_i) \approx -V^{-1} \sum_{v=1}^{V} \log p(g_v|x_i)
\end{equation}

\subsection{Coherent Persuasion Score}\label{sec:2:context}

We posit that the decreased context utilization in the case of highly popular facts \cite{mallen-etal-2023-trust, du2024context} owes to the greater likelihood attributed to the learned answer in the unconditioned output distribution (given $x_{i;q}$). Therefore, greater effort, either via parameter updates or contextual knowledge, is required to shift this output distribution to match the new answer provided in the context ($a_c$) in the conditioned instance (given $x_{i;q,c}$, Figure \ref{fig:triplets}b). This is further exacerbated when there is a superposition of competing answers (Figure \ref{fig:triplets}c). The amount of effort that must be enacted on the model parameters to output $a_c$ can be approximated by the loss, which also reflects the LM's perplexity to the output. We also measure the observed magnitude of this shift in output distribution via our novel Coherent Persuasion ($CP$) score. This score quantifies that actual efficacy of the context in shifting an LM's output distribution.

Previously, persuasion scores were introduced to assess to what degree context changes an LM's answer \cite{du2024context}. However, the existing persuasion score focuses on the first token of the single generated answer, which has been shown to be insufficient to represent the entire generated sequence \cite{wang2024firsttoken}. There also exists the problem of uncertainty during the generation of a single forward pass, which may also stem from syntactical uncertainties, which are irrelevant to our use case.

To overcome the LM's brittleness, we propose the novel Coherent Persuasion ($CP$) score. We incorporate a multiple-sample approach and semantically group the samples to create answer distributions for both $x_{i;q}$ and $x_{i;q,c}$ and compare the two. To create answer distributions, we adapt the generation and semantic grouping process as in \citet{kuhn2023semantic} into the $CP$ score.

To calculate the $CP$ score of our context $c_i$, we gather two lists of answers, $Y_{i;q}$ and $Y_{i;q,c}$, with two different inputs, which correspond to the question only input $x_{i;q}$, and the question with context input $x_{i;q,c}$. 
Then, we create the semantically similar groups $G_{i;q}=[g_1, g_2, ..., g_r, ..., g_R]$ and $G_{i;q,c}=[g_1, g_2,..., g_u, ...,g_U]$ from $Y_{i;q}$ which has $R$ groups of generated answers and $Y_{i;q,c}$ which has $U$ groups of generated answers, respectively. The $CP$ score is then obtained by averaging the divergence from the probability distribution $p$ of $G_{i;q,c}$ to $p$ of $G_{i;q}$. In detail, the output probability distribution of $g_r$ and $g_u$, is obtained as:
\begin{equation}~\label{eq:prob_pers}
    p_{g_w} = \frac{1}{W} \sum_{w=1}^{W} p_{y_{w}}
\end{equation}
where $W\in\{R,U\}$ and $p_{y_{w}}$ is the averaged softmax probability distribution of all the tokens in the answer $a_{w}$. Our final proposed $CP$ score can be acquired by Equation \ref{eq:rp_score}.
\begin{equation}~\label{eq:rp_score}
    CP(c_i) = \frac{1}{|R|\times|U|} \sum_{r=1}^{R} \sum_{u=1}^{U} KL(p_{g_r}, p_{g_u})
\end{equation}
As the $CP$ score approximates the distance between the probability distributions of $Y_{i;q,c}$ and $Y_{i;q}$, it tells us how much the LM's probability distribution is swayed by the context, while ensuring that small changes in syntactic representation or mismatches from first-token probabilities are not included. It is therefore more `coherent' in that it considers (1) the \textit{entirety} of the LM's output and (2) only the \textit{semantic} divergence in output.

\section{Experiments}

\subsection{Setup}\label{sec:setup}
We use \textsc{DynamicQA} to assess the performance of three recent and similarly sized state-of-the-art LMs: Mistral-7B-Instruct-v0.1 (Mistral, \citet{jiang2023mistral}), Llama-2-7b-chat-hf (Llama-2, \citet{touvron2023llama}), and Qwen2-7B-Instruct (Qwen2, \citet{qwen2}). To minimize the effect of the confounding factors, inferences are done in zero-shot manner, and, to obtain better generation results, instruction-tuned LMs are chosen. To obtain the model's parametric knowledge ($y_{i;q}$), we first query the model for each $q_i$ without any additional context. We then query the model provided two forms of context: one is the unperturbed context $c_o$ and the other is the unseen replacement $c_c$. In the case of disputable instances, the choice of $c_o$ and $c_c$ is arbitrary, as both options are equally likely. For each query, we obtain the accuracy and semantic entropy (\S\ref{sec:2:intra}). We calculate the $CP$ score for each $q_i$ across all queries (\S\ref{sec:2:context}). We calculate the accuracy of each model on each fact type as \[acc = \frac{\sum_{i=1}^{N}RougeL(a_i,y_i)>0.3}{N}\] where $a_i\in\{a_{c},a_{o}\}$, when context is provided, and is otherwise $a_o$. Given the inherent subjectivity of questions within the disputable portion of the dataset, we do not calculate accuracy for that partition without the context provided. For generating samples, we set $K=10$ and calculate accuracy values using the most likely response as determined via greedy search.

\begin{table*}[]
\resizebox{\textwidth}{!}{

\begin{tabular}{l|rr|rrrrrr} 
\toprule
                       & \multicolumn{1}{c}{\multirow{2}{*}{\textbf{\# of Questions}}} & \multicolumn{1}{c|}{\multirow{2}{*}{\textbf{\# of Instances}}} & \multicolumn{3}{c}{\textbf{\% of Stubborn Instances}}                                 & \multicolumn{3}{c}{\textbf{\% of Persuaded Instances}}                                 \\
                       & \multicolumn{1}{c}{}                                          & \multicolumn{1}{c|}{}                                          & \multicolumn{1}{l}{Llama-2} & \multicolumn{1}{l}{Mistral} & \multicolumn{1}{l}{Qwen2} & \multicolumn{1}{l}{Llama-2} & \multicolumn{1}{l}{Mistral} & \multicolumn{1}{l}{Qwen2}  \\ 
\midrule
\textbf{Static}        & 2500                                                          & 5000                                                           & 6.16\%             & 5.44\%          & 6.92\%                  & \textbf{78.44\%}                     & \textbf{70.52\%}                     & \textbf{61.48\%}                    \\
\textbf{Temporal}   & 2495                                                          & 4990                                                           & \textbf{9.38\%}                & \textbf{7.01\%}            & \textbf{7.54\%}                    & 60.96\%                     & 51.62\%                     & 44.81\%                    \\
\textbf{Disputable} & 694                                                           & 1388                                                           & 9.36\%                      & 6.48\%                      & 7.35\%                    & 63.83\%                     & 62.53\%                     & 59.51\%                    \\
\bottomrule
\end{tabular}

}
\caption{The number of collected questions and instances in \textsc{DynamicQA} (\S\ref{sec:dataset}) for each fact type. We also report general model behaviour (i.e. percentage of persuaded instances given context), as further described in \S \ref{sec:ana}.}
\label{tab:per_results}
\end{table*}

\begin{table*}[]
\resizebox{\textwidth}{!}{%
\begin{tabular}{ll|rrr|rrr|rrr} 
\toprule
                                     &              & \multicolumn{3}{c|}{\textbf{Accuracy ($\uparrow$)}}                                                                           & \multicolumn{3}{c|}{\textbf{Semantic Entropy ($\downarrow$)}}                                                                                                                    & \multicolumn{3}{c}{\textbf{Coherent Persuasion  Score ($\uparrow$)}}                                                                          \\ 
\cmidrule(l){3-11}
                                     &              & \multicolumn{1}{c}{\textbf{Llama-2}} & \multicolumn{1}{c}{\textbf{Mistral}} & \multicolumn{1}{c|}{\textbf{Qwen2}} & \multicolumn{1}{c}{\textbf{Llama-2}} & \multicolumn{1}{c}{\textbf{Mistral}} & \multicolumn{1}{c|}{\textbf{Qwen2}}                                             & \multicolumn{1}{c}{\textbf{Llama-2}} & \multicolumn{1}{c}{\textbf{Mistral}} & \multicolumn{1}{c}{\textbf{Qwen2}}  \\ 
\midrule
\multirow{2}{*}{\textbf{Static}}     & with context & \textbf{0.8476}                      & \textbf{0.7644}     & \textbf{0.6814}                              & 15.5663                              & 11.7557                              & 10.5875                                                               & \multirow{2}{*}{\textbf{6.8665}}     & \multirow{2}{*}{\textbf{5.8550}}              & \multirow{2}{*}{\textbf{4.1567}}             \\
                                     & w/o context  & \textbf{0.1306}                    & \textbf{0.0902}                               & \textbf{0.1244}       & 17.7064                     & \textbf{11.4943}                              & 10.4271                                                       &                                      &                                      &                                     \\ 
\midrule
\multirow{2}{*}{\textbf{Temporal}}   & with context & 0.6619                      & 0.5677                               & 0.5040                              & \textbf{15.3947}                              & \textbf{10.7685}                              & \textbf{10.5264}                                                                & \multirow{2}{*}{6.5941}     & \multirow{2}{*}{5.6314}              & \multirow{2}{*}{3.9551}             \\
                                     & w/o context  & 0.1036                      & 0.0719                               & 0.0866                              & \textbf{17.3518}                              & 11.7410                              & 11.0875                                    &                                      &                                      &                                     \\ 
\midrule
\multirow{2}{*}{\textbf{Disputable}} & with context & 0.6455                      & 0.6253                               & 0.5937                              & 16.5803                              & 11.6632                              & 10.9627                                                                & \multirow{2}{*}{5.6027}     & \multirow{2}{*}{4.1147}              & \multirow{2}{*}{3.3955}             \\
                                     & w/o context  & -                                    & -                                    & -                                   & 18.9694                              & 12.4214                              & \textbf{10.3957} &                                      &                                      &                                     \\
\bottomrule
\end{tabular}
}
\caption{The average accuracy, Semantic Entropy ($SE$; \S\ref{sec:2:intra}) and Coherent Persuasion ($CP$; \S\ref{sec:2:context}) score of our models. We bold the best values per column, with and without context. Given the inherent subjectivity of the Disputable facts, we do not show accuracy without context.}
\label{tab:gen_results}
\end{table*}

\subsection{Analysis}\label{sec:ana}
We first look into general differences in model performance on static and dynamic facts, by comparing changes in accuracy, entropy and persuasion.

We identify the difficulty of updating dynamic facts by identifying two model behaviours of interest: \textbf{Persuaded} instances are instances where the model is persuaded by the provided context, meaning that $y_{i;q} \neq a_c$ but $y_{i;q,c} = a_c$. \textbf{Stubborn} instances are instances where the model is impassive to the provided context, meaning $y_{i;q} = y_{i;q,c}$ and $y_{i;q} \neq a_c$. These are exceptional instances of RAG context usage failure. We compare the obtained $CP$ and $SE$ scores across the entire dataset and also highlight the behaviour of the persuaded and stubborn instances (\S\ref{sec:results:general}). To further explain the difficulty of retrieval-guided model updates, we observe the differences in loss (\S\ref{sec:obstacle}).

To identify factors contributing to persuasion (or RAG context usage success), we look at three characteristics of interest on the temporal dataset: semantic entropy, temporality (the number of edits), and popularity (approximated by pageviews). We analyse the Pearson correlation between these three values and our $CP$ score to identify the strongest correlates of persuasion (\S\ref{sec:interaction1}).  To further understand the influence of the various potential predictors of an instance's persuasion, we implement a logistic regression model on stubborn and persuaded instances. We take as dependent variables the popularity of the subject and object, the number of edits to the fact, and the semantic entropy before and after context is provided (\S\ref{sec:predictor}). We standardise each parameter with a $z$-score transformation for interpretable comparison.

\section{Results \& Discussion}\label{sec:results}

\subsection{General Performance}\label{sec:results:general}

Table \ref{tab:per_results} shows the percentage of stubborn and persuaded instances and Table \ref{tab:gen_results} shows the performance of our three investigated models across the three partitions of our dataset with and without additional context. Semantic Entropy ($SE$) reflects the semantic variety of model output, which can indicate internal memory conflict (\S\ref{sec:2:intra}). Lower $SE$ can be interpreted as more consistent, and less conflicted, output. On the other hand, our Coherent Persuasion ($CP$) score quantifies the actual efficacy of the context in shifting an LM's output distribution (\S\ref{sec:2:context}).

Accuracy is typically greatest with Llama-2 across all partitions. We also see that Llama-2 has the greatest degree of persuasion (in both persuasion score and \% persuaded) across all partitions. However, Llama-2 also shows the greatest entropy across all instances. This suggests \textit{Llama-2 is typically the most effective at utilising context, yet also shows the greatest intra-memory conflict}. Typically, entropy decreases with context; however, Mistral and Qwen2 show a slight increase in entropy for the Static partition of the dataset; this likely owes to the introduction of intra-memory conflict for instances with typically no competing objects. Furthermore, we typically see the \textit{greatest accuracy and persuasion (bolded) on the Static partition}. In contrast, we typically see the \textit{lowest entropy (bolded) on the Temporal dataset.} This is typically when context is provided, suggesting that the introduction of context reduces the exhibited intra-memory conflict for temporal facts. 

Our results are somewhat similar to contemporary work \cite{fierro2024mulan}, who also find the greatest accuracy with Llama-2; however, they also find the highest confidence with this model, though they presumably use the output of softmax layer which is more vulnerable to syntactic uncertainties \cite{kuhn2023semantic}. Furthermore, they find reduced update accuracy for Llama-2 relative to other models across mutable and immutable facts. In contrast, we surprisingly find that manipulations to static, immutable facts are most easily accepted by the model in context. In contrast, the temporal and disputable facts, which are expected to have greater variability in the model's training dataset, and thus higher intra-memory conflict, have a greater proportion of stubborn instances across all models. We expect this stems from our differences in dataset collection, as \citet{fierro2024mulan} determine mutability via relationship type, whereas we verify the `mutability' of our facts with our proxy measures. Ultimately, our results raise concerns for the efficacy of context usage in retrieval-augmented text generation, as \textit{we find that the most commonly adapted facts (dynamic facts) are the most difficult to update with context}.

We now investigate potential factors contributing to context usage (`persuasion') and neglect (`stubbornness'). For the main paper, we present results for Llama-2, due to its high performance, on the temporal partition of the dataset, as this partition allows for easy comparison of the effect of dynamicity on model performance. We provide all the additional graphs for the remaining models and partitions in Appendix \ref{app:extramodels}.

\begin{figure}[!t]
    \centering
    \resizebox{\columnwidth}{!}{\includegraphics[]{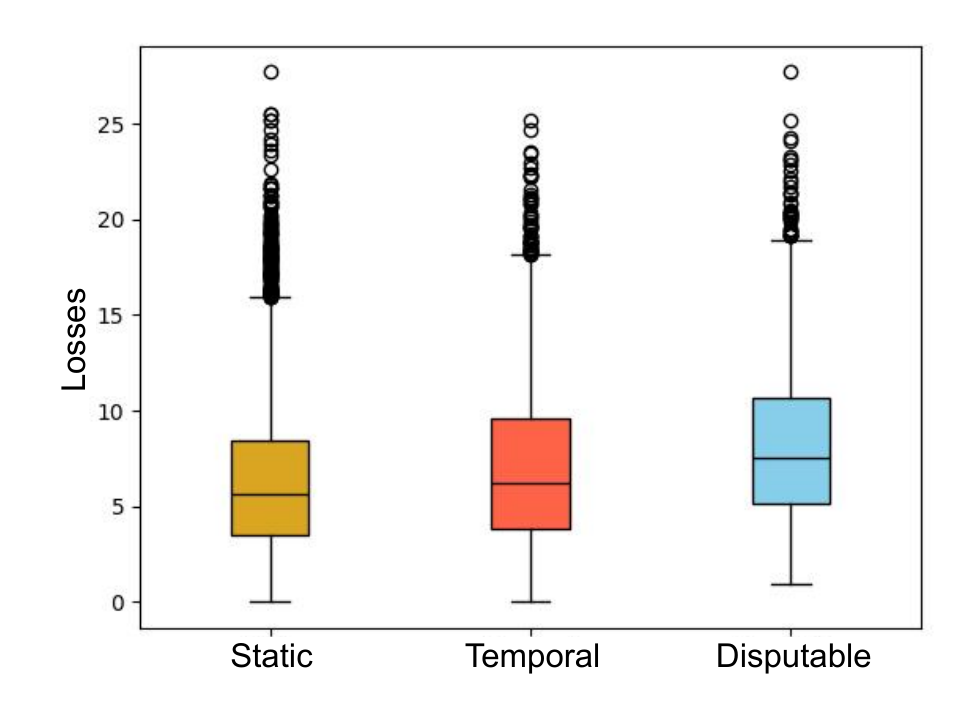}}
    \caption{The distribution of loss ($\mathcal{L}(a_c,x_{i;q})$) for generating a particular answer ($a_c$) given only the question ($x_{i;q}$) for each partition of the dataset.}
    \label{fig:loss_boxplot}
\end{figure}

\begin{figure*}[!t]
    \centering
    \resizebox{\textwidth}{!}{\includegraphics[]{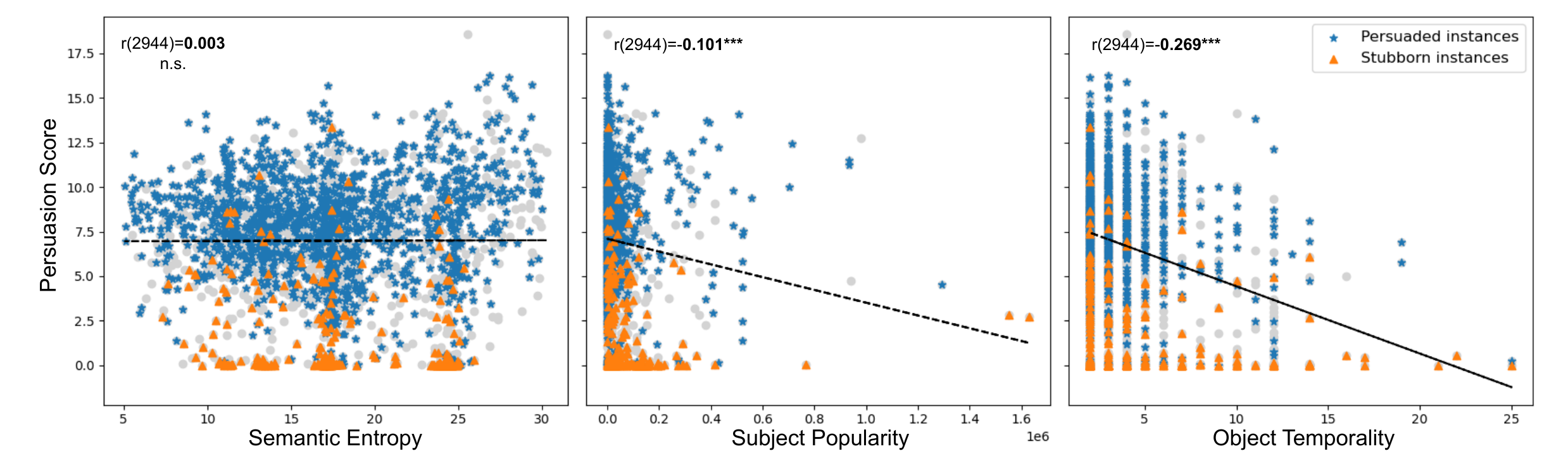}}
    \caption{The instance-level relationship between Coherent Persuasion score (\S \ref{sec:2:context}) and 3 possible factors that impact persuasion: semantic entropy, subject popularity and object temporality, alongside their Pearson correlation scores. Temporality shows the strongest relationship with persuasion. We also highlight two behaviours of interest: persuaded and stubborn instances.}
    \label{fig:scatterplot}
\end{figure*}

\subsection{Obstacles to Persuasion}\label{sec:obstacle}
To further investigate why dynamic facts show a greater proportion of stubborn instances yet a smaller divergence in output distributions, we look at the loss when we force the target answer ($a_c$) for the contextless input $x_{i;q}$. The loss reflects the likelihood of an output given the model's trained parameters. A higher loss, thereby, also indicates greater change required to steer the LM to output the target answer. Thus, we expect higher loss on dynamic facts. 
Figure \ref{fig:loss_boxplot} shows the distribution of loss for each partition of the dataset. We construct the input only with the question and calculate the losses for all the tokens in the target answer ($a_c$). The losses from the tokens in the target answer are averaged for each question. In Figure \ref{fig:loss_boxplot}, we observe that losses for temporal and disputable questions are greater than the loss for static questions; the significance is confirmed with a Welch's $t$-test ($p=9.2e-64$, $p=3.4e-18$). Therefore, while static questions show a greater $CP$ score in Table \ref{tab:gen_results}, we find \textit{there is more effort involved in updating the LM's parameters for dynamic facts, which may be unachievable with only additional context}.

\subsection{Interactors with Persuasion}\label{sec:interaction1}

We now analyse the interaction between the persuasion ($CP$) and semantic entropy ($SE$) scores, subject popularity, and object temporality in Figure \ref{fig:scatterplot} and highlight the persuaded and stubborn instances. In general, we can see that stubborn instances (orange) have lower $CP$ scores than persuaded (blue) and generic (i.e. all remaining) instances (gray). We do not see a particular correlation between the $CP$ score and $SE$ ($r(2494) = 0.003,p>.05$): meaning, \textit{instances with high initial entropy are not necessarily more likely to be persuaded to match the given context}. The distribution of persuaded instances and generic instances are identical, and there is a strong overlap with the distribution of stubborn instances. Therefore, it is difficult to determine if a retrieved context will be used by the model based on semantic entropy.

While popularity has been previously reported to negatively impact persuasion \cite{du2024context,mallen-etal-2023-trust}, popular instances are not overly represented among the stubborn instances, and the strength of the correlation ($r(2494) = 0.101,p=4e-7$), though statistically significant, is still weak. In contrast, temporality scores show the strongest correlation with $CP$ ($r(2944)=-0.27,p=1.6e-42$), suggesting \textit{there is a stronger relation between persuasion and object temporality than between persuasion and subject popularity or semantic entropy}. 

Furthermore, given the strong differences in the distributions in Figure \ref{fig:scatterplot}, we can see that semantic entropy does not indicate epistemic uncertainty (which is approximated by subject popularity) nor aleatoric uncertainty (approximated by temporality) (though semantic entropy was previously used as a measure of aleatoric uncertainty in other work \cite{gao-etal-2024-spuq}). This suggests that other factors may contribute to semantic entropy, outside of exclusively intra-memory conflict. While knowledge conflicts are a straightforward way to elicit meaningful semantic inconsistencies, they can come from other effects of model training (i.e., overlaps of objects in model memory space, spurious signals, common word co-occurrences), which may be picked up by the semantic entropy measure in addition to intra-memory conflict.

\subsection{Predictors of Persuasion}\label{sec:predictor}
In our previous investigations, we see that the semantic entropy of a contextless question is not a meaningful indicator of model persuasion, though we do see stronger relationships for popularity and temporality. Furthermore, dataset partitions with anticipated increase in intra-memory conflict (i.e., temporal and disputable data) show lower persuasion than static facts. Therefore, we assess the strength of each relation to persuasion itself. We show the results of our logistic regression test in Table \ref{tab:LM}. Though we do not find a very strong fit to the data, given its volatility ($R^2=0.139$), we do find several significant predictors. \textit{The strongest predictor of persuasion is the number of edits of a fact, where it shows an inverse relationship to a fact's persuasion}. This aligns with the behaviour we have seen in our comparison of temporal and static facts in Table \ref{tab:per_results}. Furthermore, while object popularity ($o_{pop}$) significantly affects a fact's persuasion score (previously shown by \citet{du2024context}) the magnitude of this effect is smaller than that of the number of edits of a fact. These findings typically hold across other models (see Appendix \ref{app:extramodels}), where we show that $SE_q$ typically shows the weakest relation to model persuasion, and the number of edits to have one of the strongest relationships.
Taken together, we can see that \textit{fact dynamicity, or intra-memory conflict, plays a bigger role in eliciting knowledge conflicts than fact popularity} \cite{mallen-etal-2023-trust, du2024context}. We also find \textit{an inverse effect of intra-memory conflict on an LM's susceptibility to persuasion for a given instance}. Facts that change regularly are less likely to be updated with context-retrieval, yet facts that never change are easily persuaded. 

\begin{table}[]
\centering
\small
\begin{tabular}{@{}lll@{}}
\toprule
\textbf{Predictor}  & $\boldsymbol{\widehat{\beta}}$ & \textbf{$\mathbf{p}$-value}         \\ \midrule
Intercept  &  0.87($\pm0.008$) & $p<$2e-16* \\ \midrule
\# edits   & -0.08($\pm0.008$)& $p<$2e-16* \\
$o_{pop}$ & -0.05($\pm0.008$)&  5.07e-9*          \\
$s_{pop}$ & -0.03($\pm0.008$)&2.04e-5*     \\
$SE_{c}$ & -0.04($\pm0.008$)& 1.43e-6* \\
$SE_{q}$  &  0.002($\pm0.008$)& 0.843             \\ \bottomrule
\end{tabular}

\caption{The estimated coefficients ($\hat{\beta}$) and $p$-values of the linear model predicting the persuasion score. ($R^2=0.1386$). $\hat{\beta}$ values reflect the magnitude of each predictor's effect on the dependent variable, the persuasion score, and the $p$-value denotes the statistical significance of the effect.}
\label{tab:LM}
\end{table}

\section{Conclusion}
We investigate, for the first time, the effect of intra-memory conflict on context adapation, using multiple natural causes of intra-memory conflict (i.e., fact `dynamicity'). In this, we introduce a Coherent Persuasion score to measure the persuasiveness of the given context with consideration of the entire semantic output. Furthermore, we release \textsc{DynamicQA}, a QA dataset with realistic dynamic and static questions, (i.e., facts that naturally vary and those that do not) alongside additional context and realistic replacements for use in intra-memory and counter-memory conflict studies. We then evaluate three SOTA LMs on their performance on static, temporal and disputable facts. We find that, surprisingly, static facts are the most easily updated with additional context, in comparison to temporal and disputable facts. Furthermore, our extensive analyses show that the number of unique presentations of the fact on Wikidata and Wikipedia (i.e., number of edits) has an inverse correlation with a model's propensity to adopt updates. We also find large model-level differences: Llama-2 is most easily persuaded. Over a variety of analyses, we find that uncertainty alone does not consistently indicate persuasion, requiring other approaches to update model knowledge than retrieval-augmentation in low-certainty domains.

\section*{Limitations}\label{lim}
In our paper, we do not investigate the impact of model size on knowledge conflicts, as we were computationally limited to smaller models of a size of 7B parameters. We examine several different model architectures within the model size considered. We hypothesise that the effects observed might be less pronounced for larger models. We did not include models smaller than 7B, as the ones we tested showed poor performance on the QA task in initial experiments. Other 7B parameter models we investigated showed very low accuracy, even with provided context, which is why we chose to exclude them (Falcon-7B \cite{falcon40b}, Gemma-7B \cite{gemmateam2024gemma}).

We have an imbalanced distribution of partitions in our dataset, owing to the difficulty of identifying disputable questions. We did our best to ensure our disputable dataset includes limited misinformation, though there is the possibility that some few instances made it through, given the difficulty of the task (also reflected in the relatively low annotator agreement). We also only use one possible measure of uncertainty. There has been a swarth of recent uncertainty and semantic consistency measures introduced for text generation \cite{fadeeva2023lmpolygraph}, out of which we chose the most popular one. We acknowledge relatively weak correlation values and $R^2$ reported in our paper; we owe these values to our use of realistic, rather than synthetic, retrieved contexts and distributions of popularity and temporality.

\section*{Acknowledgements}
$\begin{array}{l}\includegraphics[width=1cm]{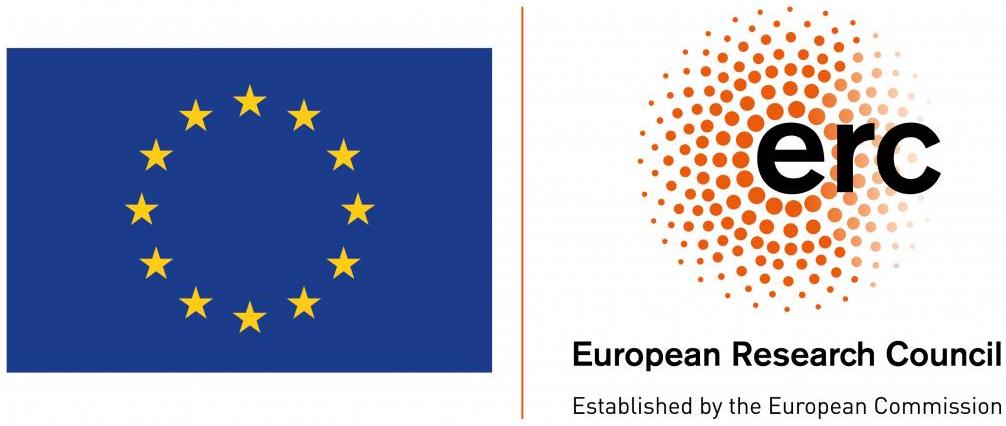} \end{array}$ 
This research was co-funded by the European Union (ERC, ExplainYourself, 101077481), by the Pioneer Centre for AI, DNRF grant number P1, as well as by The Villum Synergy Programme. We are also funded by the Villum and Velux Foundations Algorithms, Data and Democracy (ADD) grant, as well as the ERRATUM UCPH Data+ grant. Views and opinions expressed are however those of the author(s) only and do not necessarily reflect those of the European Union or the European Research Council. Neither the European Union nor the granting authority can be held responsible for them. We also thank Lovisa Hagström for her assistance in running experiments and Nadav Borenstein and Pietro Tropeano for their careful proof-reading.

\bibliography{custom}

\begin{thebibliography}{37}
\expandafter\ifx\csname natexlab\endcsname\relax\def\natexlab#1{#1}\fi

\bibitem[{Almazrouei et~al.(2023)Almazrouei, Alobeidli, Alshamsi, Cappelli, Cojocaru, Debbah, Étienne Goffinet, Hesslow, Launay, Malartic, Mazzotta, Noune, Pannier, and Penedo}]{falcon40b}
Ebtesam Almazrouei, Hamza Alobeidli, Abdulaziz Alshamsi, Alessandro Cappelli, Ruxandra Cojocaru, Mérouane Debbah, Étienne Goffinet, Daniel Hesslow, Julien Launay, Quentin Malartic, Daniele Mazzotta, Badreddine Noune, Baptiste Pannier, and Guilherme Penedo. 2023.
\newblock \href {http://arxiv.org/abs/2311.16867} {{The Falcon Series of Open Language Models}}.

\bibitem[{Chang et~al.(2024)Chang, Park, Ye, Yang, Seo, Chang, and Seo}]{chang2024largelanguagemodelsacquire}
Hoyeon Chang, Jinho Park, Seonghyeon Ye, Sohee Yang, Youngkyung Seo, Du-Seong Chang, and Minjoon Seo. 2024.
\newblock \href {http://arxiv.org/abs/2406.11813} {{How Do Large Language Models Acquire Factual Knowledge During Pretraining?}}

\bibitem[{Chen et~al.(2022)Chen, Zhang, and Choi}]{Chen2022RichKS}
Hung-Ting Chen, Michael~J.Q. Zhang, and Eunsol Choi. 2022.
\newblock \href {https://api.semanticscholar.org/CorpusID:253107178} {{Rich Knowledge Sources Bring Complex Knowledge Conflicts: Recalibrating Models to Reflect Conflicting Evidence}}.
\newblock In \emph{Conference on Empirical Methods in Natural Language Processing}.

\bibitem[{Du et~al.(2024)Du, Sn{\ae}bjarnarson, Stoehr, White, Schein, and Cotterell}]{du2024context}
Kevin Du, V{\'e}steinn Sn{\ae}bjarnarson, Niklas Stoehr, Jennifer White, Aaron Schein, and Ryan Cotterell. 2024.
\newblock \href {https://doi.org/10.18653/v1/2024.acl-long.714} {{Context versus Prior Knowledge in Language Models}}.
\newblock In \emph{Proceedings of the 62nd Annual Meeting of the Association for Computational Linguistics (Volume 1: Long Papers)}, pages 13211--13235, Bangkok, Thailand. Association for Computational Linguistics.

\bibitem[{Fadeeva et~al.(2023)Fadeeva, Vashurin, Tsvigun, Vazhentsev, Petrakov, Fedyanin, Vasilev, Goncharova, Panchenko, Panov, Baldwin, and Shelmanov}]{fadeeva2023lmpolygraph}
Ekaterina Fadeeva, Roman Vashurin, Akim Tsvigun, Artem Vazhentsev, Sergey Petrakov, Kirill Fedyanin, Daniil Vasilev, Elizaveta Goncharova, Alexander Panchenko, Maxim Panov, Timothy Baldwin, and Artem Shelmanov. 2023.
\newblock \href {https://doi.org/10.18653/v1/2023.emnlp-demo.41} {{LM}-polygraph: Uncertainty estimation for language models}.
\newblock In \emph{Proceedings of the 2023 Conference on Empirical Methods in Natural Language Processing: System Demonstrations}, pages 446--461, Singapore. Association for Computational Linguistics.

\bibitem[{Fierro et~al.(2024)Fierro, Garneau, Bugliarello, Kementchedjhieva, and S{\o}gaard}]{fierro2024mulan}
Constanza Fierro, Nicolas Garneau, Emanuele Bugliarello, Yova Kementchedjhieva, and Anders S{\o}gaard. 2024.
\newblock \href {https://aclanthology.org/2024.naacl-short.67} {{{M}u{L}an: A Study of Fact Mutability in Language Models}}.
\newblock In \emph{Proceedings of the 2024 Conference of the North American Chapter of the Association for Computational Linguistics: Human Language Technologies (Volume 2: Short Papers)}, pages 762--771, Mexico City, Mexico. Association for Computational Linguistics.

\bibitem[{Gao et~al.(2024)Gao, Zhang, Mouatadid, and Das}]{gao-etal-2024-spuq}
Xiang Gao, Jiaxin Zhang, Lalla Mouatadid, and Kamalika Das. 2024.
\newblock \href {https://aclanthology.org/2024.eacl-long.143} {{{SPUQ}: Perturbation-Based Uncertainty Quantification for Large Language Models}}.
\newblock In \emph{Proceedings of the 18th Conference of the European Chapter of the Association for Computational Linguistics (Volume 1: Long Papers)}, pages 2336--2346, St. Julian{'}s, Malta. Association for Computational Linguistics.

\bibitem[{Guu et~al.(2020)Guu, Lee, Tung, Pasupat, and Chang}]{gu2020-Realm}
Kelvin Guu, Kenton Lee, Zora Tung, Panupong Pasupat, and Mingwei Chang. 2020.
\newblock \href {https://proceedings.mlr.press/v119/guu20a.html} {{Retrieval Augmented Language Model Pre-Training}}.
\newblock In \emph{Proceedings of the 37th International Conference on Machine Learning}, volume 119 of \emph{Proceedings of Machine Learning Research}, pages 3929--3938. PMLR.

\bibitem[{Huang et~al.(2023)Huang, Yu, Ma, Zhong, Feng, Wang, Chen, Peng, Feng, Qin, and Liu}]{huang2023survey}
Lei Huang, Weijiang Yu, Weitao Ma, Weihong Zhong, Zhangyin Feng, Haotian Wang, Qianglong Chen, Weihua Peng, Xiaocheng Feng, Bing Qin, and Ting Liu. 2023.
\newblock \href {http://arxiv.org/abs/2311.05232} {{A Survey on Hallucination in Large Language Models: Principles, Taxonomy, Challenges, and Open Questions}}.

\bibitem[{Jang et~al.(2022)Jang, Ye, Lee, Yang, Shin, Han, Kim, and Seo}]{jang2023temporalwiki}
Joel Jang, Seonghyeon Ye, Changho Lee, Sohee Yang, Joongbo Shin, Janghoon Han, Gyeonghun Kim, and Minjoon Seo. 2022.
\newblock \href {https://doi.org/10.18653/v1/2022.emnlp-main.418} {{TemporalWiki: A Lifelong Benchmark for Training and Evaluating Ever-Evolving Language Models}}.
\newblock In \emph{Proceedings of the 2022 Conference on Empirical Methods in Natural Language Processing}, pages 6237--6250, Abu Dhabi, United Arab Emirates. Association for Computational Linguistics.

\bibitem[{Jiang et~al.(2023)Jiang, Sablayrolles, Mensch, Bamford, Chaplot, de~las Casas, Bressand, Lengyel, Lample, Saulnier, Lavaud, Lachaux, Stock, Scao, Lavril, Wang, Lacroix, and Sayed}]{jiang2023mistral}
Albert~Q. Jiang, Alexandre Sablayrolles, Arthur Mensch, Chris Bamford, Devendra~Singh Chaplot, Diego de~las Casas, Florian Bressand, Gianna Lengyel, Guillaume Lample, Lucile Saulnier, Lélio~Renard Lavaud, Marie-Anne Lachaux, Pierre Stock, Teven~Le Scao, Thibaut Lavril, Thomas Wang, Timothée Lacroix, and William~El Sayed. 2023.
\newblock \href {http://arxiv.org/abs/2310.06825} {{Mistral 7B}}.

\bibitem[{Kendall and Gal(2017)}]{kendallgall2017}
Alex Kendall and Yarin Gal. 2017.
\newblock \href {https://proceedings.neurips.cc/paper_files/paper/2017/file/2650d6089a6d640c5e85b2b88265dc2b-Paper.pdf} {What uncertainties do we need in bayesian deep learning for computer vision?}
\newblock In \emph{Advances in Neural Information Processing Systems}, volume~30. Curran Associates, Inc.

\bibitem[{Kuhn et~al.(2023)Kuhn, Gal, and Farquhar}]{kuhn2023semantic}
Lorenz Kuhn, Yarin Gal, and Sebastian Farquhar. 2023.
\newblock \href {https://openreview.net/forum?id=VD-AYtP0dve} {{Semantic Uncertainty: Linguistic Invariances for Uncertainty Estimation in Natural Language Generation}}.
\newblock In \emph{The Eleventh International Conference on Learning Representations}.

\bibitem[{Longpre et~al.(2021)Longpre, Perisetla, Chen, Ramesh, DuBois, and Singh}]{longpre-etal-2021-entity}
Shayne Longpre, Kartik Perisetla, Anthony Chen, Nikhil Ramesh, Chris DuBois, and Sameer Singh. 2021.
\newblock \href {https://doi.org/10.18653/v1/2021.emnlp-main.565} {{Entity-Based Knowledge Conflicts in Question Answering}}.
\newblock In \emph{Proceedings of the 2021 Conference on Empirical Methods in Natural Language Processing}, pages 7052--7063, Online and Punta Cana, Dominican Republic. Association for Computational Linguistics.

\bibitem[{Mallen et~al.(2023)Mallen, Asai, Zhong, Das, Khashabi, and Hajishirzi}]{mallen-etal-2023-trust}
Alex Mallen, Akari Asai, Victor Zhong, Rajarshi Das, Daniel Khashabi, and Hannaneh Hajishirzi. 2023.
\newblock \href {https://doi.org/10.18653/v1/2023.acl-long.546} {{When Not to Trust Language Models: Investigating Effectiveness of Parametric and Non-Parametric Memories}}.
\newblock In \emph{Proceedings of the 61st Annual Meeting of the Association for Computational Linguistics (Volume 1: Long Papers)}, pages 9802--9822, Toronto, Canada. Association for Computational Linguistics.

\bibitem[{Margatina et~al.(2023)Margatina, Wang, Vyas, Anna~John, Benajiba, and Ballesteros}]{margatina2023dynamic}
Katerina Margatina, Shuai Wang, Yogarshi Vyas, Neha Anna~John, Yassine Benajiba, and Miguel Ballesteros. 2023.
\newblock \href {https://doi.org/10.18653/v1/2023.eacl-main.211} {{Dynamic Benchmarking of Masked Language Models on Temporal Concept Drift with Multiple Views}}.
\newblock In \emph{Proceedings of the 17th Conference of the European Chapter of the Association for Computational Linguistics}, pages 2881--2898, Dubrovnik, Croatia. Association for Computational Linguistics.

\bibitem[{Neeman et~al.(2023)Neeman, Aharoni, Honovich, Choshen, Szpektor, and Abend}]{neeman-etal-2023-disentqa}
Ella Neeman, Roee Aharoni, Or~Honovich, Leshem Choshen, Idan Szpektor, and Omri Abend. 2023.
\newblock \href {https://doi.org/10.18653/v1/2023.acl-long.559} {{{D}isent{QA}: Disentangling Parametric and Contextual Knowledge with Counterfactual Question Answering}}.
\newblock In \emph{Proceedings of the 61st Annual Meeting of the Association for Computational Linguistics (Volume 1: Long Papers)}, pages 10056--10070, Toronto, Canada. Association for Computational Linguistics.

\bibitem[{Ni et~al.(2024)Ni, Bi, Guo, and Cheng}]{ni2024llmsneedretrievalaugmentation}
Shiyu Ni, Keping Bi, Jiafeng Guo, and Xueqi Cheng. 2024.
\newblock \href {https://doi.org/10.18653/v1/2024.findings-acl.675} {{When Do {LLM}s Need Retrieval Augmentation? Mitigating {LLM}s{'} Overconfidence Helps Retrieval Augmentation}}.
\newblock In \emph{Findings of the Association for Computational Linguistics ACL 2024}, pages 11375--11388, Bangkok, Thailand and virtual meeting. Association for Computational Linguistics.

\bibitem[{Pan et~al.(2023)Pan, Pan, Chen, Nakov, Kan, and Wang}]{pan2023risk}
Yikang Pan, Liangming Pan, Wenhu Chen, Preslav Nakov, Min-Yen Kan, and William Wang. 2023.
\newblock \href {https://doi.org/10.18653/v1/2023.findings-emnlp.97} {{On the Risk of Misinformation Pollution with Large Language Models}}.
\newblock In \emph{Findings of the Association for Computational Linguistics: EMNLP 2023}, pages 1389--1403, Singapore. Association for Computational Linguistics.

\bibitem[{Pezeshkpour(2023)}]{pezeshkpour2023measuring}
Pouya Pezeshkpour. 2023.
\newblock \href {https://doi.org/10.1109/ICMLA58977.2023.00122} {{Measuring and Modifying Factual Knowledge in Large Language Models}}.
\newblock \emph{2023 International Conference on Machine Learning and Applications (ICMLA)}, pages 831--838.

\bibitem[{Rad and Barbosa(2012)}]{wikicontroversystudy}
Hoda~Sepehri Rad and Denilson Barbosa. 2012.
\newblock \href {https://doi.org/10.1145/2462932.2462942} {{Identifying controversial articles in Wikipedia: a comparative study}}.
\newblock In \emph{Proceedings of the Eighth Annual International Symposium on Wikis and Open Collaboration}, WikiSym '12, New York, NY, USA. Association for Computing Machinery.

\bibitem[{Shanahan et~al.(2023)Shanahan, McDonell, and Reynolds}]{Shanahan_McDonell_Reynolds_2023}
Murray Shanahan, Kyle McDonell, and Laria Reynolds. 2023.
\newblock \href {https://doi.org/10.1038/s41586-023-06647-8} {{Role play with large language models}}.
\newblock \emph{Nature}, 623(7987):493–498.

\bibitem[{Song et~al.(2024)Song, Park, Hwang, Yun, Joe, Gwon, and Yoon}]{song-etal-2024-entity}
Jongyoon Song, Nohil Park, Bongkyu Hwang, Jaewoong Yun, Seongho Joe, Youngjune Gwon, and Sungroh Yoon. 2024.
\newblock \href {https://aclanthology.org/2024.eacl-long.55} {{Entity-level Factual Adaptiveness of Fine-tuning based Abstractive Summarization Models}}.
\newblock In \emph{Proceedings of the 18th Conference of the European Chapter of the Association for Computational Linguistics (Volume 1: Long Papers)}, pages 915--929, St. Julian{'}s, Malta. Association for Computational Linguistics.

\bibitem[{Sumi et~al.(2011)Sumi, Yasseri, Rung, Kornai, and Kertesz}]{editwars}
R.~Sumi, T.~Yasseri, A.~Rung, A.~Kornai, and J.~Kertesz. 2011.
\newblock \href {https://doi.org/10.1109/PASSAT/SocialCom.2011.47} {{Edit Wars in Wikipedia}}.
\newblock In \emph{{2011 IEEE Third International Conference on Privacy, Security, Risk and Trust (PASSAT) / 2011 IEEE Third International Conference on Social Computing (SocialCom)}}, pages 724--727, Los Alamitos, CA, USA. IEEE Computer Society.

\bibitem[{Team et~al.(2024)Team, Mesnard, Hardin, Dadashi, Bhupatiraju, Pathak, Sifre, Rivière, Kale, Love, Tafti, Hussenot, Sessa, Chowdhery, Roberts, Barua, Botev, Castro-Ros, Slone, Héliou, Tacchetti, Bulanova, Paterson, Tsai, Shahriari, Lan, Choquette-Choo, Crepy, Cer, Ippolito, Reid, Buchatskaya, Ni, Noland, Yan, Tucker, Muraru, Rozhdestvenskiy, Michalewski, Tenney, Grishchenko, Austin, Keeling, Labanowski, Lespiau, Stanway, Brennan, Chen, Ferret, Chiu, Mao-Jones, Lee, Yu, Millican, Sjoesund, Lee, Dixon, Reid, Mikuła, Wirth, Sharman, Chinaev, Thain, Bachem, Chang, Wahltinez, Bailey, Michel, Yotov, Chaabouni, Comanescu, Jana, Anil, McIlroy, Liu, Mullins, Smith, Borgeaud, Girgin, Douglas, Pandya, Shakeri, De, Klimenko, Hennigan, Feinberg, Stokowiec, hui Chen, Ahmed, Gong, Warkentin, Peran, Giang, Farabet, Vinyals, Dean, Kavukcuoglu, Hassabis, Ghahramani, Eck, Barral, Pereira, Collins, Joulin, Fiedel, Senter, Andreev, and Kenealy}]{gemmateam2024gemma}
Gemma Team, Thomas Mesnard, Cassidy Hardin, Robert Dadashi, Surya Bhupatiraju, Shreya Pathak, Laurent Sifre, Morgane Rivière, Mihir~Sanjay Kale, Juliette Love, Pouya Tafti, Léonard Hussenot, Pier~Giuseppe Sessa, Aakanksha Chowdhery, Adam Roberts, Aditya Barua, Alex Botev, Alex Castro-Ros, Ambrose Slone, Amélie Héliou, Andrea Tacchetti, Anna Bulanova, Antonia Paterson, Beth Tsai, Bobak Shahriari, Charline~Le Lan, Christopher~A. Choquette-Choo, Clément Crepy, Daniel Cer, Daphne Ippolito, David Reid, Elena Buchatskaya, Eric Ni, Eric Noland, Geng Yan, George Tucker, George-Christian Muraru, Grigory Rozhdestvenskiy, Henryk Michalewski, Ian Tenney, Ivan Grishchenko, Jacob Austin, James Keeling, Jane Labanowski, Jean-Baptiste Lespiau, Jeff Stanway, Jenny Brennan, Jeremy Chen, Johan Ferret, Justin Chiu, Justin Mao-Jones, Katherine Lee, Kathy Yu, Katie Millican, Lars~Lowe Sjoesund, Lisa Lee, Lucas Dixon, Machel Reid, Maciej Mikuła, Mateo Wirth, Michael Sharman, Nikolai Chinaev, Nithum Thain, Olivier Bachem,
  Oscar Chang, Oscar Wahltinez, Paige Bailey, Paul Michel, Petko Yotov, Rahma Chaabouni, Ramona Comanescu, Reena Jana, Rohan Anil, Ross McIlroy, Ruibo Liu, Ryan Mullins, Samuel~L Smith, Sebastian Borgeaud, Sertan Girgin, Sholto Douglas, Shree Pandya, Siamak Shakeri, Soham De, Ted Klimenko, Tom Hennigan, Vlad Feinberg, Wojciech Stokowiec, Yu~hui Chen, Zafarali Ahmed, Zhitao Gong, Tris Warkentin, Ludovic Peran, Minh Giang, Clément Farabet, Oriol Vinyals, Jeff Dean, Koray Kavukcuoglu, Demis Hassabis, Zoubin Ghahramani, Douglas Eck, Joelle Barral, Fernando Pereira, Eli Collins, Armand Joulin, Noah Fiedel, Evan Senter, Alek Andreev, and Kathleen Kenealy. 2024.
\newblock \href {http://arxiv.org/abs/2403.08295} {{Gemma: Open Models Based on Gemini Research and Technology}}.

\bibitem[{Touvron et~al.(2023)Touvron, Martin, Stone, Albert, Almahairi, Babaei, Bashlykov, Batra, Bhargava, Bhosale, Bikel, Blecher, Ferrer, Chen, Cucurull, Esiobu, Fernandes, Fu, Fu, Fuller, Gao, Goswami, Goyal, Hartshorn, Hosseini, Hou, Inan, Kardas, Kerkez, Khabsa, Kloumann, Korenev, Koura, Lachaux, Lavril, Lee, Liskovich, Lu, Mao, Martinet, Mihaylov, Mishra, Molybog, Nie, Poulton, Reizenstein, Rungta, Saladi, Schelten, Silva, Smith, Subramanian, Tan, Tang, Taylor, Williams, Kuan, Xu, Yan, Zarov, Zhang, Fan, Kambadur, Narang, Rodriguez, Stojnic, Edunov, and Scialom}]{touvron2023llama}
Hugo Touvron, Louis Martin, Kevin Stone, Peter Albert, Amjad Almahairi, Yasmine Babaei, Nikolay Bashlykov, Soumya Batra, Prajjwal Bhargava, Shruti Bhosale, Dan Bikel, Lukas Blecher, Cristian~Canton Ferrer, Moya Chen, Guillem Cucurull, David Esiobu, Jude Fernandes, Jeremy Fu, Wenyin Fu, Brian Fuller, Cynthia Gao, Vedanuj Goswami, Naman Goyal, Anthony Hartshorn, Saghar Hosseini, Rui Hou, Hakan Inan, Marcin Kardas, Viktor Kerkez, Madian Khabsa, Isabel Kloumann, Artem Korenev, Punit~Singh Koura, Marie-Anne Lachaux, Thibaut Lavril, Jenya Lee, Diana Liskovich, Yinghai Lu, Yuning Mao, Xavier Martinet, Todor Mihaylov, Pushkar Mishra, Igor Molybog, Yixin Nie, Andrew Poulton, Jeremy Reizenstein, Rashi Rungta, Kalyan Saladi, Alan Schelten, Ruan Silva, Eric~Michael Smith, Ranjan Subramanian, Xiaoqing~Ellen Tan, Binh Tang, Ross Taylor, Adina Williams, Jian~Xiang Kuan, Puxin Xu, Zheng Yan, Iliyan Zarov, Yuchen Zhang, Angela Fan, Melanie Kambadur, Sharan Narang, Aurelien Rodriguez, Robert Stojnic, Sergey Edunov, and Thomas
  Scialom. 2023.
\newblock \href {http://arxiv.org/abs/2307.09288} {{Llama 2: Open Foundation and Fine-Tuned Chat Models}}.

\bibitem[{Vuong et~al.(2008)Vuong, Lim, Sun, Le, Lauw, and Chang}]{controversy2}
Ba-Quy Vuong, Ee-Peng Lim, Aixin Sun, Minh-Tam Le, Hady~Wirawan Lauw, and Kuiyu Chang. 2008.
\newblock \href {https://doi.org/10.1145/1341531.1341556} {{On ranking controversies in wikipedia: models and evaluation}}.
\newblock In \emph{Proceedings of the 2008 International Conference on Web Search and Data Mining}, WSDM '08, page 171–182, New York, NY, USA. Association for Computing Machinery.

\bibitem[{Wan et~al.(2024)Wan, Wallace, and Klein}]{wan2024evidence}
Alexander Wan, Eric Wallace, and Dan Klein. 2024.
\newblock \href {https://doi.org/10.18653/v1/2024.acl-long.403} {{What Evidence Do Language Models Find Convincing?}}
\newblock In \emph{Proceedings of the 62nd Annual Meeting of the Association for Computational Linguistics (Volume 1: Long Papers)}, pages 7468--7484, Bangkok, Thailand. Association for Computational Linguistics.

\bibitem[{Wang et~al.(2024)Wang, Ma, Hu, Weber-Genzel, R{\"o}ttger, Kreuter, Hovy, and Plank}]{wang2024firsttoken}
Xinpeng Wang, Bolei Ma, Chengzhi Hu, Leon Weber-Genzel, Paul R{\"o}ttger, Frauke Kreuter, Dirk Hovy, and Barbara Plank. 2024.
\newblock \href {https://doi.org/10.18653/v1/2024.findings-acl.441} {{{``}My Answer is {C}{''}: First-Token Probabilities Do Not Match Text Answers in Instruction-Tuned Language Models}}.
\newblock In \emph{Findings of the Association for Computational Linguistics ACL 2024}, pages 7407--7416, Bangkok, Thailand and virtual meeting. Association for Computational Linguistics.

\bibitem[{Xie et~al.(2024)Xie, Zhang, Chen, Lou, and Su}]{xie2024sloth}
Jian Xie, Kai Zhang, Jiangjie Chen, Renze Lou, and Yu~Su. 2024.
\newblock \href {https://openreview.net/forum?id=auKAUJZMO6} {{Adaptive Chameleon or Stubborn Sloth: Revealing the Behavior of Large Language Models in Knowledge Conflicts}}.
\newblock In \emph{The Twelfth International Conference on Learning Representations}.

\bibitem[{Xu et~al.(2024{\natexlab{a}})Xu, Lin, Yang, Zhang, Shi, Zhang, Fang, Xu, and Qiu}]{xu2024earth}
Rongwu Xu, Brian Lin, Shujian Yang, Tianqi Zhang, Weiyan Shi, Tianwei Zhang, Zhixuan Fang, Wei Xu, and Han Qiu. 2024{\natexlab{a}}.
\newblock \href {https://doi.org/10.18653/v1/2024.acl-long.858} {{The Earth is Flat because...: Investigating {LLM}s{'} Belief towards Misinformation via Persuasive Conversation}}.
\newblock In \emph{Proceedings of the 62nd Annual Meeting of the Association for Computational Linguistics (Volume 1: Long Papers)}, pages 16259--16303, Bangkok, Thailand. Association for Computational Linguistics.

\bibitem[{Xu et~al.(2024{\natexlab{b}})Xu, Qi, Wang, Wang, Zhang, and Xu}]{xu2024knowledge}
Rongwu Xu, Zehan Qi, Cunxiang Wang, Hongru Wang, Yue Zhang, and Wei Xu. 2024{\natexlab{b}}.
\newblock \href {http://arxiv.org/abs/2403.08319} {{Knowledge Conflicts for LLMs: A Survey}}.

\bibitem[{Yang et~al.(2024)Yang, Yang, Hui, Zheng, Yu, Zhou, Li, Li, Liu, Huang, Dong, Wei, Lin, Tang, Wang, Yang, Tu, Zhang, Ma, Yang, Xu, Zhou, Bai, He, Lin, Dang, Lu, Chen, Yang, Li, Xue, Ni, Zhang, Wang, Peng, Men, Gao, Lin, Wang, Bai, Tan, Zhu, Li, Liu, Ge, Deng, Zhou, Ren, Zhang, Wei, Ren, Liu, Fan, Yao, Zhang, Wan, Chu, Liu, Cui, Zhang, Guo, and Fan}]{qwen2}
An~Yang, Baosong Yang, Binyuan Hui, Bo~Zheng, Bowen Yu, Chang Zhou, Chengpeng Li, Chengyuan Li, Dayiheng Liu, Fei Huang, Guanting Dong, Haoran Wei, Huan Lin, Jialong Tang, Jialin Wang, Jian Yang, Jianhong Tu, Jianwei Zhang, Jianxin Ma, Jianxin Yang, Jin Xu, Jingren Zhou, Jinze Bai, Jinzheng He, Junyang Lin, Kai Dang, Keming Lu, Keqin Chen, Kexin Yang, Mei Li, Mingfeng Xue, Na~Ni, Pei Zhang, Peng Wang, Ru~Peng, Rui Men, Ruize Gao, Runji Lin, Shijie Wang, Shuai Bai, Sinan Tan, Tianhang Zhu, Tianhao Li, Tianyu Liu, Wenbin Ge, Xiaodong Deng, Xiaohuan Zhou, Xingzhang Ren, Xinyu Zhang, Xipin Wei, Xuancheng Ren, Xuejing Liu, Yang Fan, Yang Yao, Yichang Zhang, Yu~Wan, Yunfei Chu, Yuqiong Liu, Zeyu Cui, Zhenru Zhang, Zhifang Guo, and Zhihao Fan. 2024.
\newblock \href {http://arxiv.org/abs/2407.10671} {{Qwen2 Technical Report}}.

\bibitem[{Yu et~al.(2024)Yu, Atanasova, and Augenstein}]{yu2024revealingparametricknowledgelanguage}
Haeun Yu, Pepa Atanasova, and Isabelle Augenstein. 2024.
\newblock \href {https://doi.org/10.18653/v1/2024.acl-long.444} {Revealing the parametric knowledge of language models: A unified framework for attribution methods}.
\newblock In \emph{Proceedings of the 62nd Annual Meeting of the Association for Computational Linguistics (Volume 1: Long Papers)}, pages 8173--8186, Bangkok, Thailand. Association for Computational Linguistics.

\bibitem[{Yu et~al.(2023)Yu, Merullo, and Pavlick}]{yu-etal-2023-characterizing}
Qinan Yu, Jack Merullo, and Ellie Pavlick. 2023.
\newblock \href {https://doi.org/10.18653/v1/2023.emnlp-main.615} {{Characterizing Mechanisms for Factual Recall in Language Models}}.
\newblock In \emph{Proceedings of the 2023 Conference on Empirical Methods in Natural Language Processing}, pages 9924--9959, Singapore. Association for Computational Linguistics.

\bibitem[{Zhang and Choi(2023)}]{zhang2024mitigating}
Michael Zhang and Eunsol Choi. 2023.
\newblock \href {https://doi.org/10.18653/v1/2023.emnlp-main.879} {{Mitigating Temporal Misalignment by Discarding Outdated Facts}}.
\newblock In \emph{Proceedings of the 2023 Conference on Empirical Methods in Natural Language Processing}, pages 14213--14226, Singapore. Association for Computational Linguistics.

\bibitem[{Årup Nielsen(2017)}]{nielsen2017wembedder}
Finn Årup Nielsen. 2017.
\newblock \href {http://arxiv.org/abs/1710.04099} {{Wembedder: Wikidata entity embedding web service}}.

\end{thebibliography}
\bibliographystyle{acl_natbib}

\appendix

\section{Additional Methodology Details}
\label{sec:appendix:a}

\subsection{Annotation}
Given the source of the dataset and the potential for vandalism in the dataset, three early-career researchers (PhD, postdoc) and volunteers from the research group annotated all questions, using the following guidelines: For cases of multiple questions for the same piece of context, the most specific question is kept, or the question that best suits the context and two potential objects. If both questions are identical, the first is kept and the second is discarded. If one question is better than the other, keep that question and discard the other. Annotators are allowed to look information up online to make a decision. The annotator is given 5 options: \begin{enumerate}
    \item \textbf{Accept}. The question is related to the context. The two potential objects are possible answers to the question. Neither of the potential objects contain vandalism or obvious misinformation. 
    \item \textbf{Change}. If both questions are incorrect or insufficient, the first question is edited. It must be rewritten so that most objects are acceptable answers to the question based on the context. If the question cannot be rewritten in a meaningful way, it should be discarded (i.e., if both words are adjectives). To ensure the text is specific enough, most extra information might be added. The annotator should also remove all "according to the text"s as this would not work well without context.
    \item  \textbf{Discard}. If there is no dispute for a question, it should be discarded. This could be if the two objects are two names for the same person (e.g., Kanye, Ye), two different spellings for the same word, or two synonyms (publications, writings). One exception is hypernyms (e.g., Danish, Scandinavian). We discard questions with insufficient context (e.g., `later jesus t.'). We remove fixed typos and vandalism and clear misinformation (e.g., the objects `china' and `ukraine' to the question `Which state agency is responsible for managing the Chernobyl exclusion zone and has offices on the site?'). 
    \item \textbf{Misinformation}. If clearly both answers cannot be correct, we make it as misinformation for further review before discarding.
    \item \textbf{Incomplete}. If fixes are needed to the context, but it is otherwise acceptable, we will process these extra datapoints to ensure it is usable.
\end{enumerate}

Two annotators annotated every instance. In the case of disagreement, a third annotator annotated the instance. We kept instances marked with a majority `accept' or `accept' and `change'. Annotators were the group of early-career NLP researchers (PhD, postdoc) and volunteers from the research group.

\begin{table*}[]
\resizebox{\textwidth}{!}{%
\begin{tabular}{@{}ll@{}}
\toprule
                             & \multicolumn{1}{c}{\textbf{Prompt $\mathbf{P}$}}                                                                                                                                                                                                                                                                                                                                                                                                                                                                                                                                                                                                                                                                                                                                                                                   \\ \midrule
\textbf{Input $[P;q_i]$}     & \begin{tabular}[c]{@{}l@{}}"System": You'll be given a question and a context about the article and answer it with a one word. Answer the [Question],\\ "User": This article is about Titanic. Who was the producer of Titanic?\end{tabular}                                                                                                                                                                                                                                                                                                                                                                                                                                                                                                                                                                              \\ \midrule
\textbf{Input $[P;c_i;q_i]$} & \begin{tabular}[c]{@{}l@{}}"System": You'll be given a question and a context about the article and answer it with a one word. Answer the [Question],\\ "User": This article is about Titanic. Titanic is a 1997 American epic romantic disaster film \\ directed, written, produced, and co-edited by James Cameron.\\ Incorporating both historical and fictionalized aspects, it is based on accounts of the sinking of RMS Titanic in 1912.\\ Leonardo DiCaprio and Kate Winslet star as members of different social classes who fall in love during the ship's maiden voyage.\\ The film also features an ensemble cast of Billy Zane, Kathy Bates, Frances Fisher, \\Gloria Stuart, Bernard Hill, Jonathan Hyde, Victor Garber, David Warner, Suzy Amis and Bill Paxton.\\ Who was the producer of Titanic?\end{tabular} \\ \bottomrule
\end{tabular}%
}
\caption{Example of a prompt according to the presence of the context $c_i$ in the input. The example here is the question that asks about the producer of the movie Titanic with the answer-specific context which mentions answer (James Cameron) in the context.}
\label{tab:appd:prompt}
\end{table*}

\begin{table*}[]
\resizebox{\textwidth}{!}{%
\begin{tabular}{@{}l@{}}
\toprule
\multicolumn{1}{c}{\textbf{Prompt}}                                                                                                                                                                                                                                                                                                                    \\ \midrule
"System": Given the [Context], Generate the [Question] whose correct answer is the [Answer]. The [Answer] is also highlighted in the [Context] with hl. \\
For example, [Context]: <hl> jainism <hl>, traditionally known as jain dharma, is an ancient indian religion. [Answer]: jainism \\ 
 Example [Question]: Which ancient indian religion is known as jain dharma? \\
"User": This article is about KFC.
[Context]: KFC responded by adding a cheap hot <hl> burger <hl> to the menu, \\ 
called a “snacker”, which is easier to eat than chicken on the bone. [Answer]: burger [Question]:                                                                               \\ \bottomrule
\end{tabular}%
}
\caption{Example of a prompt for question generation. The LM is asked to generate the question about KFC. The expected answer of the question is ``burger".}
\label{tab:prompt_qgen}
\end{table*}

\subsection{Question Generation}\label{app:qgen}

For automatic generation of the questions for disputable facts, we feed the context and answer to the Meta-Llama-3-8B-Instruct model with the prompt that contains the example of the expected behaviour. The prompt used for the question generation is presented in Table \ref{tab:prompt_qgen}.
During the generation, the hyperparameter top\_p is set to 0.9 and the temperature is 0.6. We remove the question if the generated question contains the intended answer.

\subsection{Implementation details}\label{app:implementation}

For a NLI model that is used to determine the semantic similarity score between two generated answers (\S\ref{sec:2:intra}), we use a Deberta-large model that is finetuned on the NLI dataset called MNLI \footnote{\url{https://huggingface.co/microsoft/deberta-large-mnli}}. With two generated answers ($a_{i,k-1}$, $a_{i,k}$), an input is constructed as $[CLS] a_{i,k-1} [SEP] a_{i,k} [SEP]$. To ensure that A1, A2 entail each other, we perform two forward passes with two different inputs by changing the order of the answer within the input ($[CLS] a_{i,k-1} [SEP] a_{i,k} [SEP]$). They are regarded as semantically similar when both predictions of two inputs are `entailment'. For the grouping of semantically similar answers, the $k$-th generated answer $a_{i,k}$ from an input $x_i$ is assigned to the group $g_v$ if $a_{i,k}$ entails other answers within the $g_v$. See \citet{kuhn2023semantic} for more details.

For the semantic similarity model used in \S\ref{sec:Disputability}, we use the transformer encoder model\footnote{\url{https://huggingface.co/sentence-transformers/all-MiniLM-L6-v2}} that is trained in a contrastive manner to distinguish the similarity between two sentences.

Table \ref{tab:appd:prompt} showcases the prompts used to inference the model. For Mistral, Llama-2, we apply chat template to our input since they were trained with the chat template format. To generate, we follow sampling approach and set the maximum number of generated tokens to 20.

\section{Additional Graphs}\label{app:extramodels}
\begin{figure*}
    \resizebox{\textwidth}{!}{\includegraphics[]{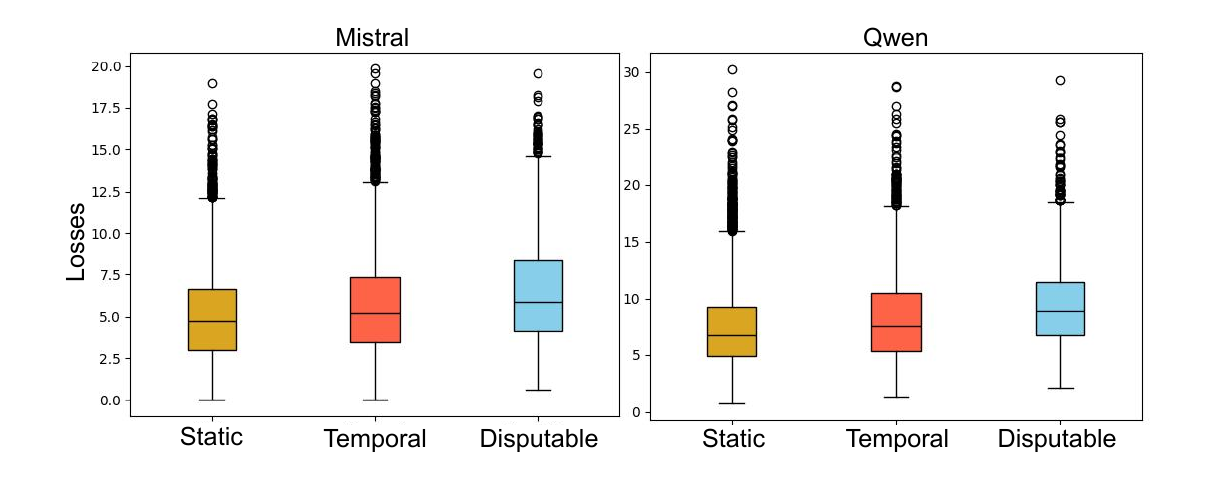}}
    \caption{The distribution of losses for each partition of the dataset on Mistral and Qwen}
    \label{fig:loss_appendix}
\end{figure*}

\begin{figure*}
    \resizebox{\textwidth}{!}{\includegraphics[]{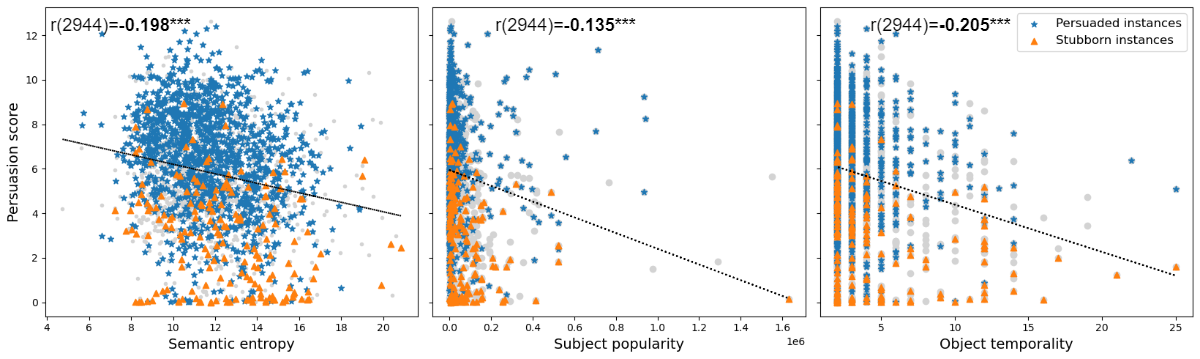}}
    \caption{The instance-level relationship between Coherent Persuasion Score and 3 possible factors that impact
persuasion: semantic entropy, subject popularity and object temporality, alongside their Pearson correlation scores.
Temporality shows the strongest relationship with persuasion. We also highlight two behaviours of interest:
persuaded and stubborn instances. These results are for \textbf{Mistral}.}
    \label{fig:mistral_scatter}
\end{figure*}

\begin{figure*}
    \resizebox{\textwidth}{!}{\includegraphics[]{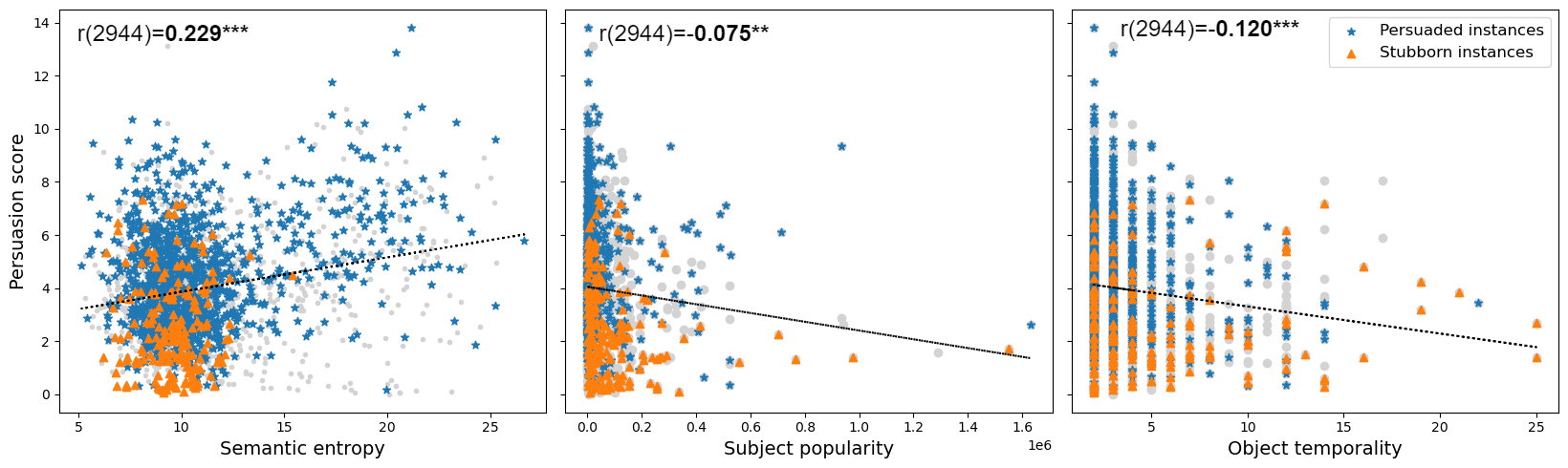}}
    \caption{The instance-level relationship between Coherent Persuasion Score and 3 possible factors that impact
persuasion: semantic entropy, subject popularity and object temporality, alongside their Pearson correlation scores.
Temporality shows the strongest relationship with persuasion. We also highlight two behaviours of interest:
persuaded and stubborn instances. These results are for \textbf{Qwen}.}
    \label{fig:qwen_scatter}
\end{figure*}

Here we present the results for additional models for the analyses in 
\S\ref{sec:results}. Figure \ref{fig:loss_appendix} shows that the analysis on the loss (\S\ref{sec:obstacle}) holds the same across the different models.  Figures \ref{fig:mistral_scatter} and \ref{fig:qwen_scatter} show that temporality has a consistently negative and relatively strong relationship with the $CP$ score across models. While semantic entropy can also be a relatively strong correlate, it changes sign between models, showing that this relationship is inconsistent; meaning uncertain models may be more or less likely to utilise context. Tables \ref{tab:mistrallm} and \ref{tab:qwenlm} show that we continually see a strong effect of number of edits as a predictor of persuasion across models.

\begin{table}[h]
\centering
\small
\begin{tabular}{@{}lll@{}}
\toprule
\textbf{Predictor}  & $\boldsymbol{\widehat{\beta}}$ & \textbf{$\mathbf{p}$-value}         \\ \midrule
Intercept  &  0.88($\pm0.01$) & $p<$2e-16* \\ \midrule
\# edits   & -0.05($\pm0.01$)& 5.92e-12* \\
$o_{pop}$ & -0.06($\pm0.01$)& 1.36e-13*          \\
$s_{pop}$ & -0.03($\pm0.01$)& 5.22e-4*         \\
$SE_{c}$ & 0.02($\pm0.01$)& 8.74e-3*     \\
$SE_{q}$  &  -0.02($\pm0.01$)& 7.97e-3*             \\ \bottomrule
\end{tabular}
\caption{Results of the Logistic Regression model on Mistral Temporal Results. ($R^2=0.116$)}
\label{tab:mistrallm}
\end{table}

\begin{table}[h]
\centering
\small
\begin{tabular}{@{}lll@{}}
\toprule
\textbf{Predictor}  & $\boldsymbol{\widehat{\beta}}$ & \textbf{$\mathbf{p}$-value}         \\ \midrule
Intercept  &  0.86($\pm0.01$) & $p<$2e-16* \\ \midrule
\# edits   & -0.07($\pm0.01$)& 4.60e-15* \\
$o_{pop}$ & -0.06($\pm0.01$)& 1.64e-12*          \\
$s_{pop}$ & -0.04($\pm0.01$)& 1.21e-4*          \\
$SE_{c}$ & 0.08($\pm0.01$)& $p<$2e-16*     \\
$SE_{q}$  &  0.03($\pm0.01$)& 1.40e-4*             \\ \bottomrule
\end{tabular}
\caption{Results of the Logistic Regression model on Qwen2 Temporal Results. ($R^2=0.197$)}
\label{tab:qwenlm}
\end{table}

\end{document}